\documentclass[journal]{IEEEtran}
\usepackage[pdftex]{graphicx}
\usepackage{epsfig}
\usepackage{epstopdf}
\usepackage{caption, color}
\usepackage{array}
\usepackage{multirow}
\usepackage{tabularx,booktabs}
\usepackage{amsmath}
\usepackage{amssymb}
\usepackage{algorithm}
\usepackage{algorithmic}
\usepackage{url}

\begin{document}

\title{Design Identification of Curve Patterns on Cultural Heritage Objects: Combining Template Matching and CNN-based Re-Ranking}

\author{Jun~Zhou,
        Yuhang~Lu,
        Kang~Zheng,~\IEEEmembership{Student Member, IEEE},
        Karen~Smith,
        Colin~Wilder,
        and Song~Wang,~\IEEEmembership{Senior Member, IEEE}
\thanks{Jun Zhou, Yuhang Lu, Kang Zheng and Song Wang are with the Department of Computer Science \& Engineering, University of South Carolina, SC, USA.
}
\thanks{Karen Smith is with the South Carolina Institute of Archaeology and Anthropology, University of South Carolina, SC, USA.}
\thanks{Colin Wilder is with the Center for Digital Humanities at the University of South Carolina, SC, USA.}
}

\maketitle

\begin{abstract}
The surfaces of many cultural heritage objects were embellished with various patterns, especially curve patterns. In practice, most of the unearthed cultural heritage objects are highly fragmented, e.g., sherds of potteries or vessels, and each of them only shows a very small portion of the underlying full design, with noise and deformations. The goal of this paper is to address the challenging problem of automatically identifying the underlying full design of curve patterns from such a sherd. Specifically, we formulate this problem as template matching: curve structure segmented from the sherd is matched to each location with each possible orientation of each known full design. In this paper, we propose a new two-stage matching algorithm, with a different matching cost in each stage. In Stage 1, we use a traditional template matching, which is highly computationally efficient, over the whole search space and identify a small set of candidate matchings. In Stage 2, we derive a new matching cost by training a dual-source Convolutional Neural Network (CNN) and apply it to re-rank the candidate matchings identified in Stage 1. We collect 600 pottery sherds with 98 full designs from the Woodland Period in Southeastern North America for experiments and the performance of the proposed algorithm is very competitive.  
\end{abstract}



\section{Introduction}
\label{sec:introduction}

\begin{figure*}[htbp]
	\begin{center}
		\includegraphics[width=1\linewidth]{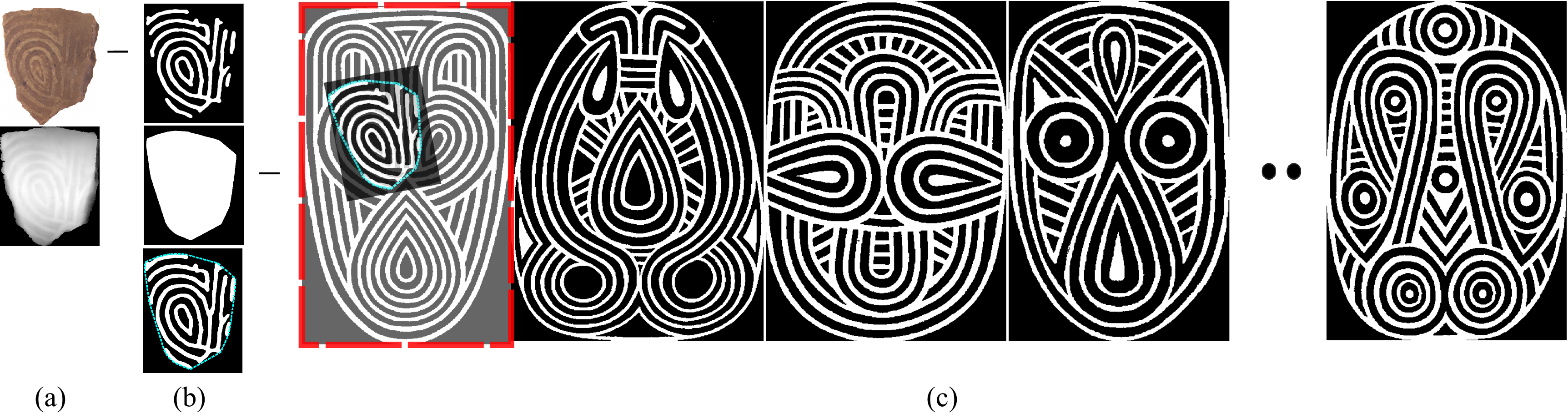}
	\end{center}
	\vspace{-0.5em}
	\caption{An illustration of identifying the underlying design for a pottery sherd. (a) A sherd's RGB and depth images. (b) The curve structure segmented from the sherd. From top to bottom are the segmented curve structure of the sherd and the mask derived from the sherd boundary, and the segmented curve structure masked by the sherd boundary. (c) A database of known designs, where the true design with the best matching is highlighted in the red box. Original design reproduced with permission, courtesy of
Frankie Snow, South Georgia State College.}
	\label{fig:1-procedure}
	\vspace{-0.5em}
\end{figure*}

{M}illions of archived cultural heritage objects such as bone, pottery, shell, wood, and cloth are very precious records in archeology -- many of these objects are embellished with various man-made patterns, especially curve patterns, and the \emph{designs} of these patterns provide important information to archaeologists. However, most of these cultural heritage objects are highly fragmented, e.g. potsherds rather than whole vessels, and each of them only show a small portion of the underlying full design of patterns. For example, Figure~\ref{fig:1-procedure}(a) shows one pottery sherd from the Woodland Period in Southeastern North America (300-600 AD), when the Native Americans had a tradition of decorating pottery by stamping carved wooden paddle on the pottery surface. The full curve pattern carved on the wooden paddle, as shown in Fig.~\ref{fig:1-procedure}(c), is the underlying design, which can be used to build chronologies, to track trade networks, and to reconstruct aspects of style and the creative process. An important problem in archeology is to automatically and quickly identify the underlying design from the partial pattern shown on the surface of a sherd~\cite{zhou2017identifying}.

In this paper, we investigate this important problem by focusing on \emph{curve} patterns, where the underlying full design and the partial pattern on the sherd are curve structures, as shown in Fig.~\ref{fig:1-procedure}. More specifically, after decades of efforts from archaeologists, many full designs have been revealed, reconstructed and archived for different periods and regions in archeology, such as the Woodland Period in Southeastern North America. This way, the design identification problem can be formulated as identifying the best matched design for a sherd by segmenting the curve structure from the sherd and then matching it against a set of known designs.  As illustrated in Fig.~\ref{fig:1-procedure}, we can match the curve structure segmented from a sherd, as shown in Fig.~\ref{fig:1-procedure}(b), against each location, with each possible orientation, of each known design, and then select the design with the lowest matching cost as the matched design. This exhaustive matching procedure identifies not only the matched design, but also the matched location and orientation on the matched design. Note that, scale transform is usually not considered in the matching since both sherds and full designs have known size in real world and their matchings are not scale invariant, i.e., two designs are considered different in archeology even if they are identical after a uniform scale transform.
\begin{figure}[htbp]
            \begin{center}
                        \includegraphics[width=\linewidth]{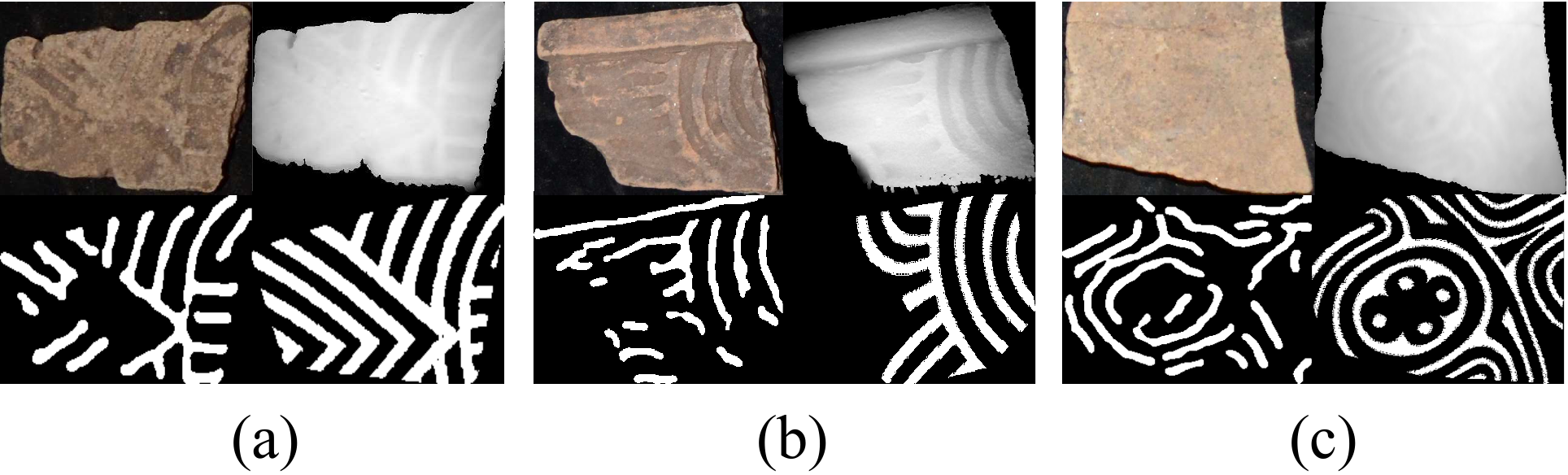}
            \end{center}
            \vspace{-0.5em}
            \caption{ An illustration of various noise and deformations in the curve structures segmented from sherds: (a) A sherd with missing curve segments due to erosion; (b) A sherd with noisy curves due to weathering and/or shallow paddle stamping on rough pottery surface; (c) A sherd with deformed curve patterns due to the drying process in making the pottery. For each case, we show the sherd's RGB image (top-left), the depth image (top-right), the segmented curve structure (bottom-left) and the corresponding potion in the underlying design (bottom-right).}
            \label{fig:2-difficultsherds}
            \vspace{-1em}
\end{figure} 
Based on this problem formulation, the key issue is then the definition of an appropriate cost in matching the curve structure segmented from a sherd to a location of a full design, with a specified orientation. This problem is nontrivial in the proposed archeology application for two reasons. First, the exhaustive matching against each possible location and orientation of each design leads to a very large search space. To prevent from an overly slow algorithm, we require the matching cost to be very efficient to compute for each possible solution in the search space. Second, compared with the underlying design, the curve structure segmented from the sherd usually contain strong noise and deformations in the drying process in making many of these objects, many years of erosion and sediment under the earth, and the imperfectness of the curve-segmentation algorithms, as illustrated by several examples in Fig.~\ref{fig:2-difficultsherds}. 

To address this problem, in this paper we propose a new two-stage matching algorithm, with a different matching cost in each stage. In Stage 1, we propose to use a classical template matching method, which is highly computational efficient, over the whole search space to identify a small set of candidate matchings on all the known designs. This simple matching cost can help efficiently reduce the search space of solutions. In Stage 2, we further derive a new matching cost by training a dual-source Convolutional Neural Network (CNN) and apply this more computational-intensive matching to re-rank the candidate matchings identified in Stage 1. Through supervised learning, we expect that various kinds of noise and deformations in the segmented curve structures, as shown in Fig.~\ref{fig:2-difficultsherds}, can be implicitly identified and suppressed in computing the CNN-based matching cost.
 
In the experiments, we collected the images of 600 pottery sherds excavated from archaeological sites associated with the Woodland period paddle-stamping tradition, together with their corresponding 98 full paddle designs, to evaluate the performance of the proposed two-stage matching algorithm. Based on the Cumulative Matching Characteristics (CMC) ranking metric, the proposed algorithm can achieve a much higher accuracy than several other existing matching algorithms that are chosen for performance comparison in the experiment. 

\section{Related Work}

Many previous works on image processing of cultural heritage objects are focused on \textbf{fragment classification and matching}, which aims to recognize sherds that are originally from the same object, such as a pottery vessel, and then assemble these sherds to reconstruct the original 3D object. For example, Durham et al.~\cite{durham1995artefact} applied a generalized Hough transform to perform artifact retrieval and then assembled them based on edge detection. Smith et al.~\cite{smith2010classification} proposed a method for thin ceramic sherd classification based on color and texture characteristics using total different geometry energies. Schurmans et al.~\cite{schurmans2002advances} studied the sherd uniformity and standardization using points or areas from a 3D model and the profile of the studied shapes. 
Different from these works, in this paper our goal is to identify the design of curve patterns on a sherd and the sherds identified with the same design are usually not from the same object, e.g., the same vessel. As a result, we could not assemble them for 3D reconstruction.

\textbf{Curve-structure matching} has been a long studied problem in computer vision and image processing. By thinning all the curve structure to one-pixel wide, many shape matching algorithms have been developed for matching curve structures. For example, Belongie et al.~\cite{belongie2001shape} proposed a shape context approach, which builds a log-polar histogram around each sampled curve point and then uses this histogram as the feature to match two curve structures. Barrow et al.~\cite{barrow1977parametric} proposed the widely used Chamfer matching algorithm by pre-computing a distance map for efficiently locating one curve structure on another curve structure. By treating curve structure as binary images, image-based matching algorithms can be directly applied here for matching curve structures. For example, Brunelli~\cite{brunelli2009template} developed a template matching method by treating one as a convolution mask over the other. In Perceptual Hash (pHash)~\cite{zauner2010implementation}, each image was coded into a 64-bit fingerprint number, which was then used as features to compare and match images. However, many of these existing matching algorithms are sensitive to noise and deformation present in the curve structures segmented from sherds.

In principle, \textbf{local feature matching}~\cite{hassaballah2016image} can also be used for handling curve-structure matching -- we can extract a set of image features, such as the Gabor~\cite{manjunath1996texture}, the local binary patterns (LBP)~\cite{ojala2002multiresolution}, the histogram of oriented gradients (HOG)~\cite{dalal2005histograms} or the scale-invariant feature transform (SIFT)~\cite{brown2007automatic}, from both the image of curve structure segmented from a sherd and the image of the design, and match them by identifying a set of corresponding features. However, most of these local feature detectors are developed for color or gray-scale images and could not accurately capture the critical local features in binary images of curve structures. Furthermore, these image feature detectors and the corresponding feature descriptors are not robust enough to the noise and deformation present in curve structures segmented from sherds.

In Stage 2 of the proposed method, we employ a dual-source CNN for matching two curve structures. \textbf{Deep neural network based image matching} have been studied by several research groups in recent years. For example, a patch-based local image matching network, called MatchNet~\cite{han2015matchnet},  was developed to jointly learn the feature representation and the matching function from image data. In~\cite{zagoruyko2015learning}, after exploring multiple neural networks, a CNN-based model called DeepCompare was developed to match a variety of images based on their appearance. In~\cite{lin2015deep}, Lin et al. designed a CNN-based model for fast image retrieval using binary hash codes. In~\cite{wang2014learning}, Wang et al. proposed a deep ranking approach by learning fine-grained image similarity. Several networks have been proposed in related applications. For example, AlexNet~\cite{krizhevsky2012imagenet} claimed victory at the LSVRC2012 classification contest. ResNet~\cite{he2016deep}, GoogleNet~\cite{43022} and VGGNet~\cite{simonyan2014very} later drew a lot of attention in large scale image recognition and object detection. However, these methods are mainly developed for matching color or gray-scale images and show degraded performance in matching binary images of curve structures with noise and deformation. 

In the later experiments, we include several of these existing matching algorithms for performance evaluation and comparison. 

\section{Proposed Method}

\begin{figure*}[htbp]
	\begin{center}
		\includegraphics[width=0.8\linewidth]{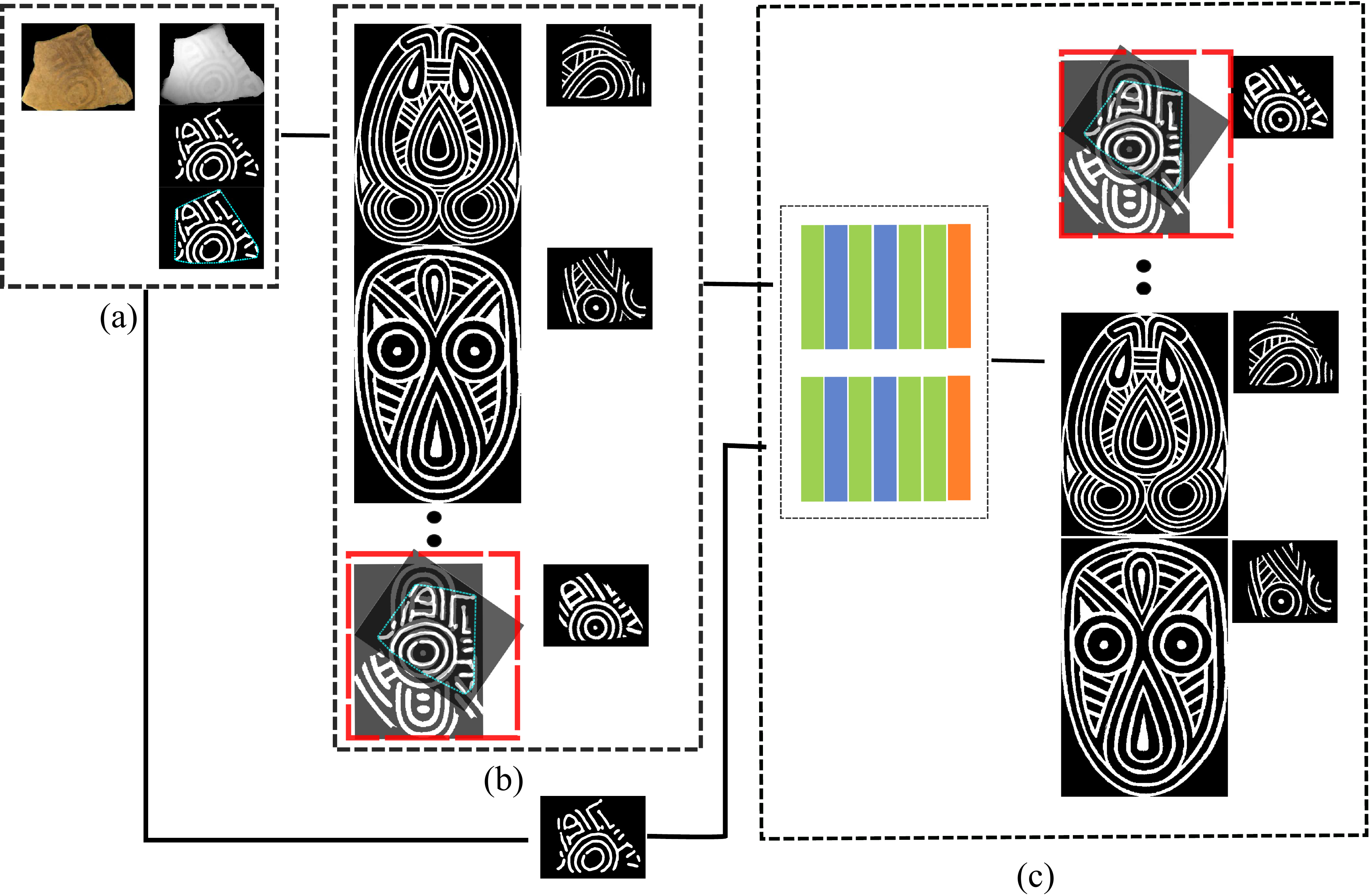}
	\end{center}
	\vspace{-0.5em}
	\caption{An illustration of the full pipeline of the proposed method. (a) Curve structure  segmentation from a sherd. (b) Stage 1: template matching with all the designs for selecting a small set of candidate matchings of the input sherd. (c) Stage 2: CNN-based re-ranking of the candidate matchings. For (b)  and (c), from top to bottom the resulting matchings are shown in ascending order with respect to their costs. Correctly matching design is shown in red box, which is ranked low in Stage 1 but ranked at the top in Stage 2. Original design reproduced with permission, courtesy of Frankie Snow, South Georgia State College.}
	\label{fig:3-pipeline}
\end{figure*}

The full pipeline of the proposed method is illustrated in Fig.~\ref{fig:3-pipeline}. 
 Given the image of a sherd, we first segment the curve structure in the form of a binary image masked by the sherd boundary. We then develop a two-stage algorithm for design identification. 1) \emph{Template matching}. In this stage, we apply a simple template matching to match the curve structure segmented from the sherd over all the locations, with all possible orientations, of all the designs and select a small set of candidate matchings. 2) \emph{CNN-based re-ranking.} In this stage, we derive a new matching cost by training a dual-source Convolutional Neural Network (CNN) and apply it to re-rank the candidate matchings for final matching designs, as well as the matching locations and orientations. In the following, we first introduce the curve structure segmentation from the image of a sherd, and then elaborate on the two stages of the proposed algorithm.
 
\subsection{Curve-Structure Segmentation from a Sherd} \label{sec:segmentation}

\begin{figure}[htbp]
 	\begin{center} 		
 		\includegraphics[width=1\linewidth]{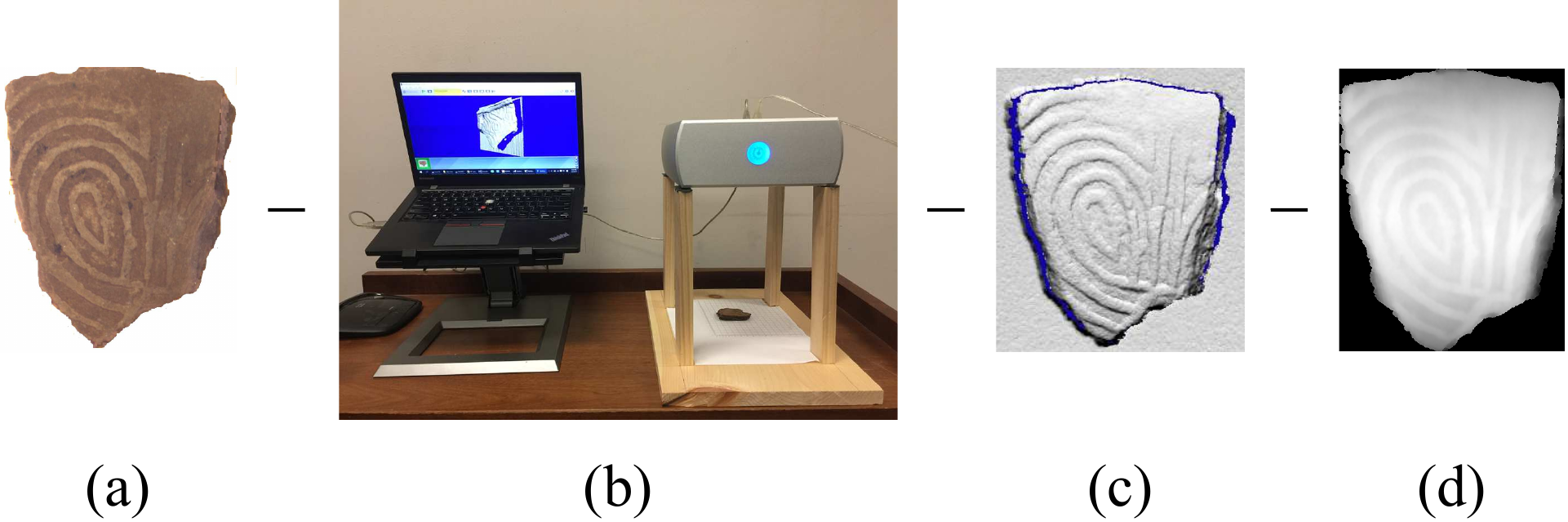}
 	\end{center}
 	\vspace{-0.5em}
 	\caption{An illustration of scanning sherds for depth images: (a) RGB image of a sherd, (b) setup of a 3D scanner, (c) 3D point cloud of the sherd surface obtained by the 3D scanner, and (d) depth image of the sherd surface: pixel intensity represents the depth value at a location.}
 	\label{fig:4-scansherd}
 	\vspace{-0.5em}
 \end{figure}

Generally speaking, extracting curve structures from the surface of a sherd is a typical low-level image segmentation problem.  However, the erosion and sediment usually make the curve structures on the sherd very weak and blurred, which substantially increases the difficulty in accurately segmenting them. In this paper, we use the excavated pottery sherds associated with the Woodland period for experiments and we found that it is very difficult to extract these curve structures from the camera-taken RGB images of these sherds. Given that these curve structures are stamped on the surfaces of pottery vessels by carved paddles, curve structures usually show larger depth than the adjacent non-curve surface. Therefore, in archeology, 3D scanners are usually applied to achieve the 3D depth image of the sherd surface, as illustrated in Fig.~\ref{fig:4-scansherd}, and the curve structures are then segmented directly from the depth image. 

\begin{figure}[htbp]
 	\begin{center} 		
 		\includegraphics[width=0.8\linewidth]{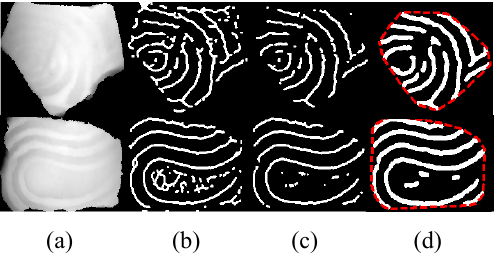}
 	\end{center}
 	\vspace{-0.5em}
 	\caption{An illustration of segmenting curve structures from sample sherds: (a) depth images of sherds, where darker pixels have larger depths. (b) FCN-extracted curve skeletons, (c) Refined curve skeletons by using a dense prediction CNN, and (d) final segmented curve structures with recovered curve width, masked by the sherd boundaries (indicated by red contours).}
 	\label{fig:5-curveextraction}
 	\vspace{-1em}
 \end{figure}

However, after buried under the earth for thousands of years, together with possible shallow stamping in making the vessel, the curve structures can still be difficult to segment even from the scanned high-resolution depth images. In our previous work~\cite{lu2018curve}, we propose a new CNN-based algorithm to more accurately and reliably segment the stamped curve structures from the depth images of the sherds, by learning and incorporating the implied curve geometry, such as curve smoothness and parallelism, in the underlying designs. Specifically, we train a Fully Convolutional Network (FCN) to detect the skeletons of the curve structures in the depth images. Then, we train a dense prediction convolutional network to identify and prune false positive skeleton pixels. Finally, we recover the curve width by a scale-adaptive thresholding algorithm to get the final segmentation of curve structures. Figure~\ref{fig:5-curveextraction} shows the sample results after each step of this algorithm. We also extract the boundary of sherd, indicated by red contours in Fig.~\ref{fig:5-curveextraction}. The sherd boundary provides a mask to exclude all the information outside the sherd boundary from matching in the later two-stage algorithm. 

It has been shown in~\cite{lu2018curve} that, this CNN-based algorithm can segment the curve structure from a sherd much more accurately than other low-level and high-level image segmentation algorithms. However, the segmented curves are far from perfect, because of the strong noise and shallow stampings on the unearthed sherds, as shown in Fig.~\ref{fig:2-difficultsherds}. In particular, many segmentation errors may occur near the sherd boundary, where the depth information is more noisy due to the erosion and sediment. Furthermore, as mentioned above, the curve structure segmented from a sherd may show deformation from its underlying design due to the drying process in making the object. In the following, we elaborate on the proposed two-stage matching algorithm that is robust to noise, errors and deformation present in the segmented curve structures.

\subsection{Stage 1:  Template Matching}

In Stage 1, we treat the design identification as a traditional template matching problem.  The curve structure segmented from a sherd is taken as a \emph{template}, which is a binary image $I_T$ with mask $M_T$. The $N$ known designs, each of which is a binary image $I_i$, $i=1,2,\cdots, N$, are taken as \emph{sources}. At location $\mathbf{p}=(x,y)$ with orientation $\theta$ on design $I_i$,  we adopt the simple template matching cost 
\begin{equation}
\phi_i(\mathbf{p},\theta)= \sum_{(x',y')\in M_T^{(\mathbf{p},\theta)}} \left (I_T^{(\mathbf{p},\theta)}(x',y')-I_i(x',y')\right )^2,
\label{eq:template-matching-cost}
\end{equation}
where $(I_T^{(\mathbf{p},\theta)}, M_T^{(\mathbf{p},\theta)})$ is the template with translation $\mathbf{p}$ and rotation angle $\theta$.  In this stage of the algorithm, we calculate the matching cost $\phi_i(\mathbf{p},\theta)$ exhaustively for the whole search space -- for design $I_i$, the translation $\mathbf{p}$ contains all the pixel locations in $I_i$ and the orientation angle $\theta$ covers all the 360 integer degrees in the range of $[0, 360^{\circ})$.  This way, just for design $I_i$, we need to calculate the matching cost $360\times S_i$ times with $S_i$ being the size of $I_i$, i.e., the number of pixels in $I_i$.  Combining all $N$ designs, the whole search space for calculating the matching cost in this stage is $360\times \sum_{i=1}^N S_i$, which can be very large when the number of designs $N$ increases and the size of the design increases. But given the high simplicity of this matching cost on the binary images, its calculation over the whole search space can be still achieved very quickly by applying fast cross-correlation of $I_T$ and $I_i$~\cite{lewis1995fast}.  

Given the noise, errors and deformation in the template $(I_T^{(\mathbf{p},\theta)}, M_T^{(\mathbf{p},\theta)})$ and the simplicity of the matching cost Eq.~(\ref{eq:template-matching-cost}), the globally optimal matching
$\arg\min_{(i,\mathbf{p},\theta)}\phi_i(\mathbf{p},\theta)$ over the whole search space may not be the desired correct matching. Therefore, in this stage we select a small set of candidate matchings with relatively low template matching costs of Eq.~(\ref{eq:template-matching-cost}), with an expectation that the desired ground-truth matching is included in these candidate matchings. The final matching and design identification will be determined by a more refined matching cost, which we will discuss later in Stage 2.

For simplicity,  for a given template, we select $K$ candidate matchings on each design $i$, $i=1,2,\cdots, N$. In selecting the $K$ candidate matchings, we need to avoid the inclusion of spatially adjacent ones, e.g., the matchings $(\mathbf{p},\theta)$ and $(\mathbf{p},\theta+1^{\circ})$ on the same design $i$ since they may show very similar matching cost. To address this problem, we employ the following ``non-minimum suppression" strategy to select the $K$ candidate matchings on the design $i$. 

\begin{figure}[htbp]
	\centering
	\includegraphics[scale=0.3]{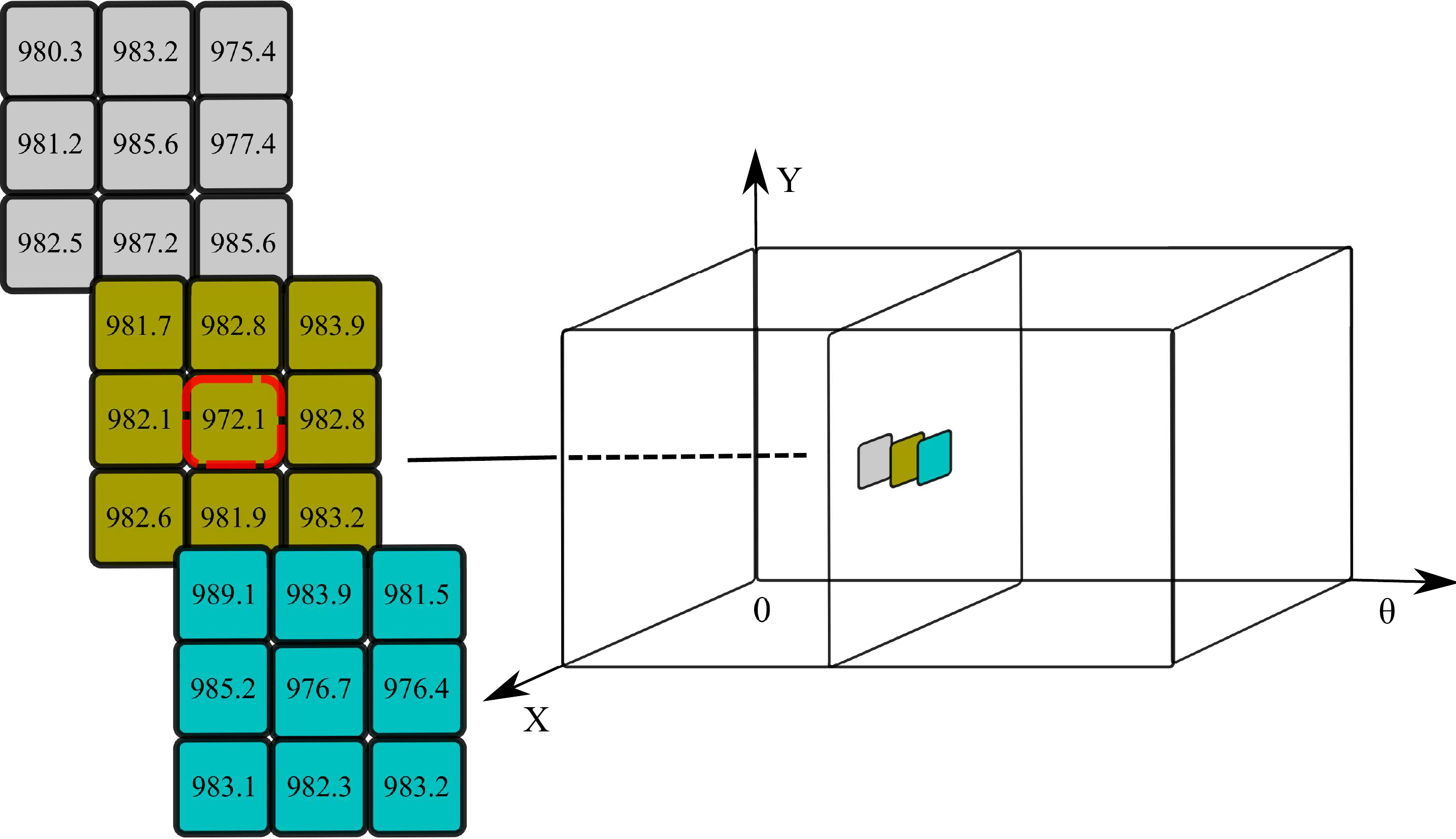}
	\vspace{+0.5em}
	\caption{3D non-minimum suppression for selecting the candidate matchings. Local minimums are defined over a $3\times 3\times 3$ window over the matrix constructed by the template matching costs computed over the whole search space. }
	\label{fig:6-nms-3D}
\end{figure}

\begin{enumerate}
	\item On the design $i$, compute the template matching cost $\phi_i(\mathbf{p},\theta)=\phi_i(x,y,\theta)$ over the whole search space. This leads to a 3D matrix with dimensions of $x$, $y$, and $\theta$.
	\item Find the local minimums of this 3D matrix, i.e., all the elements $(x,y,\theta)$ that satisfy
	$$\phi_i(x,y,\theta)\leq \phi_i(x+\Delta x,y+\Delta y,\theta+\Delta \theta)$$ with $\Delta x$, $\Delta y$, and $\Delta \theta$ take value in $\{-1, 0, 1\}$. On the design, we apply rotation symmetry, i.e., $\phi_i(x,y,\theta+360^{\circ})=\phi_i(x,y,\theta)$ for padding along the dimension of $\theta$ and the traditional float maximum padding along the dimensions of $x$ and $y$. An example is shown in Fig.~\ref{fig:6-nms-3D}.
	\item Select the $K$ of these local minimums with the lowest matching cost as the candidate matchings on design $i$. If there are less than $K$ local minimums, we simply take all the local minimums as the candidate matchings on design $i$.	
\end{enumerate}

This way, for each design $i$, we have $K$ candidate matchings\footnote{The degenerated cases of less than $K$ local minimums on a design does not make any difference to the later algorithm development.}, which we denote as $(\mathbf{p}_i^{k},\theta_i^{k})$, $k=1,2,\cdots, K$ and $i=1,2,\cdots, N$. In the following section, we introduce a CNN-based algorithm to re-rank these $N\times K$ candidate matchings for the final matching and design identification.

\subsection{Stage 2: CNN-based Re-ranking}

\begin{figure}[htbp]
\begin{center}
   
   \includegraphics[width=1\linewidth]{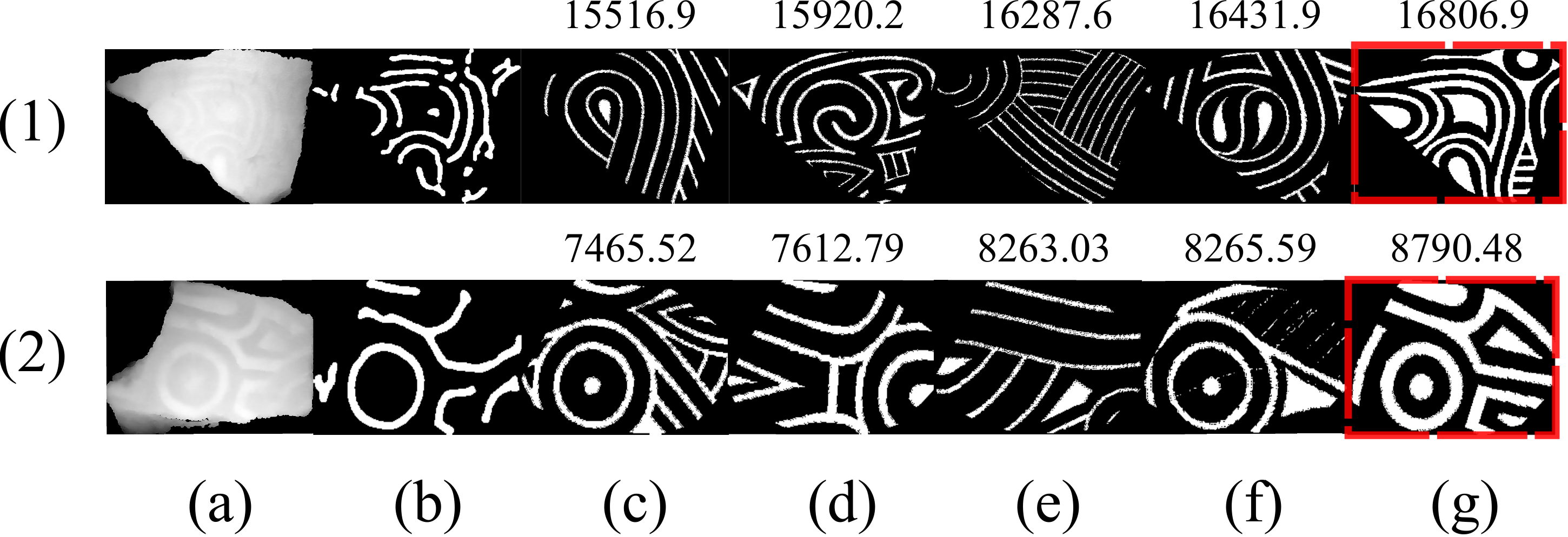}
  \vspace{-1.5em}
\end{center}
   \caption{Two examples of ranking candidate matchings using the template-matching cost of Eq.~(\ref{eq:template-matching-cost}) with each row showing an example. (a) Depth images of sherds, (b) templates, i.e., the curve structures segmented from the sherds, (c-g) top five ranked candidate matchings (from the first to the fifth) and the corresponding matching costs. For each design, only the matched portion masked by the corresponding sherd boundary is displayed. Red boxes indicate the desired true matchings.}
\label{fig:7-failurematching}
\end{figure}

For the curve structure segmented from a sherd, the template matching in Stage 1 provides us $N\times K$ candidate matchings, i.e., $K$ candidate matchings on each design. The remaining problem is to rank these $N\times K$ candidate matchings to locate the best matchings, with which we can identify the best matched designs for the input sherd. The simplest way is to directly use the template matching cost in Eq.~(\ref{eq:template-matching-cost}) to rank these candidate matchings. However, as mentioned above, the noise, errors, and deformation in the template may prevent the use of this simple template-matching cost from making the correct ranking and identifying the underlying design. As shown in Fig.~\ref{fig:7-failurematching}, a false matching in the candidates may show lower template-matching cost of Eq.~(\ref{eq:template-matching-cost}) than the true matching, i.e., the true location and orientation on the true design.

We can see that the incorrect rankings may be caused by different kinds of degradations such as noise, errors and deformation of the template. For example, the segmentation error near the sherd boundary may be high, as shown in the top row of Fig.~\ref{fig:7-failurematching}. Furthermore, the curve structure in certain area of the sherd may be totally missing due to shallow stamping in making the object, as shown in  in the bottom row of Fig.~\ref{fig:7-failurematching}. It is very difficult to find and explicitly model all the possible degradations and propose specific strategies to tackle these degradations. Instead, in this stage of the algorithm, we propose to develop a CNN-based algorithm to learn features that are robust to these degradations in a supervised way. We then use these robust features to calculate the matching cost and re-rank the candidate matchings derived in Stage 1 for final design identification.

\begin{figure}[htbp]
	\centering
	\includegraphics[scale=0.3]{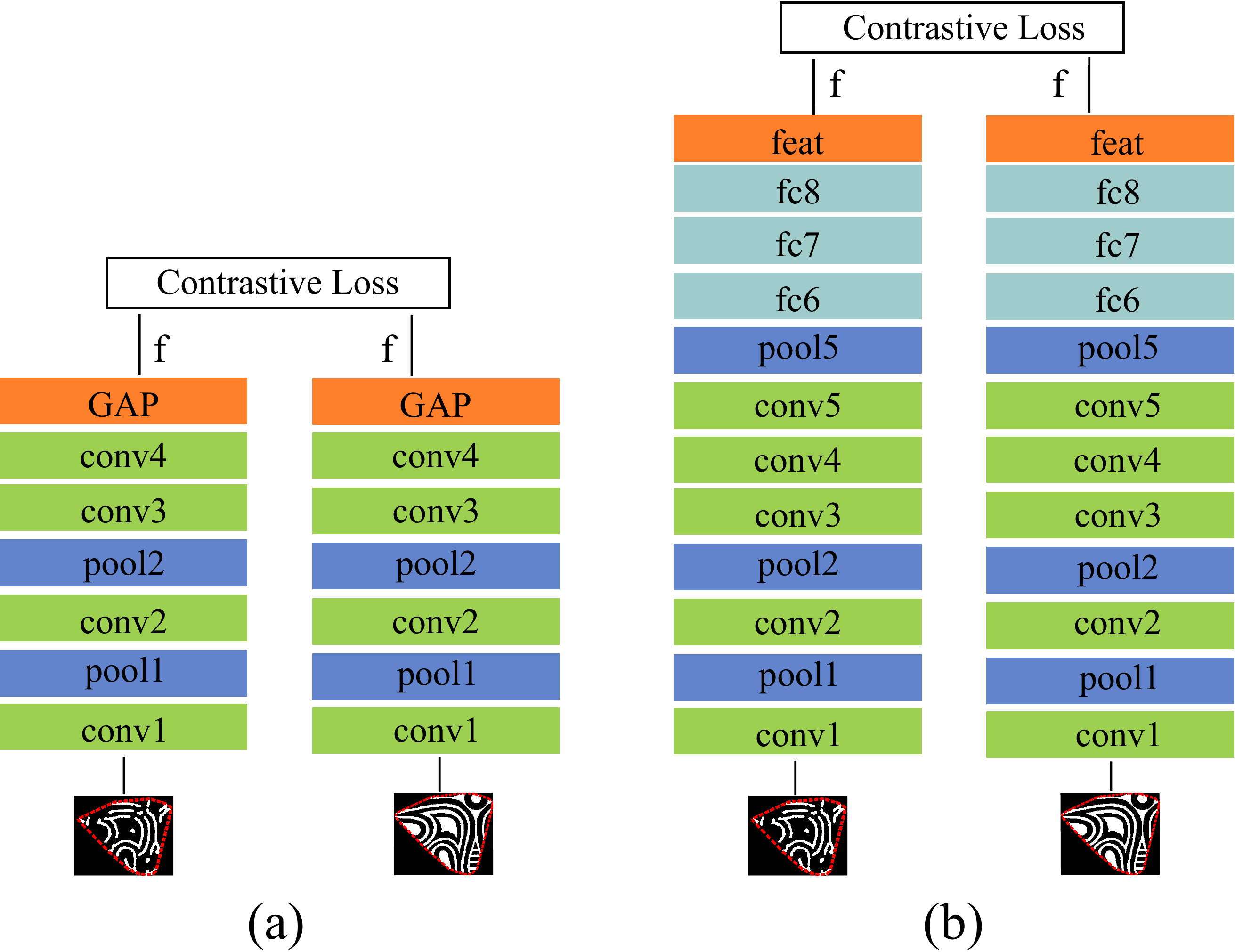}
	\caption{An illustration of the dual-source CNN architectures: (a) the proposed CNN and (b) AlexNet.}\label{fig:8-network}
	\vspace{-1em}
\end{figure}

For each candidate matching $(i, \mathbf{p}_i^{k},\theta_i^{k})$, it actually matches the curve structure segmented from the sherd, i.e., $(I_T, M_T)$ and an identical-size patch $I_i^k$ of the design $I_i$ defined by
the location and orientation of $(\mathbf{p}_i^{k},\theta_i^{k})$ and the same mask $M_T$, as illustrated in Fig.~\ref{fig:7-failurematching}(c). By setting the intensity of all the pixels outside the mask $M_T$ to be zero, all we need is a robust matching cost between binary images $I_T$ and $I_i^k$. To solve this problem, we propose a dual-source CNN with two identical sub-networks as shown in Fig.~\ref{fig:8-network}(a). These two sub-networks take $I_T$ and $I_i^k$ as the inputs, respectively. Each sub-network consists of a sequence of convolution, max pooling layers and a global average pooling layer (GAP) for feature learning, as detailed in Table~\ref{net_config}. We implement this dual-source CNN by truncating AlexNet~\cite{krizhevsky2012imagenet}, as shown in Fig.~\ref{fig:8-network}(b), to ``conv4" layer and replacing all layers after ``conv4" layer with a GAP layer. Both inputs, i.e., $I_T$ and $I_i^k$ are re-sized to $227\times227$ pixels, before being fed to the sub-network.

\begin{table}[htbp]
\caption{Configuration of each sub-network; $n, k, s, p$ stand for the number of outputs, kernel size, stride and padding size respectively.}
	\centering
	\begin{tabular}{|c |c | c|} 
		\hline
		\rule{0pt}{10pt} \textbf{Name} & \textbf{Type} & \textbf{Configuration}\\
		\hline\hline
		\rule{0pt}{9pt} GAP & GlobalAveragePooling & $k:13\times 13, s:1$ \\ \hline
		\rule{0pt}{9pt} conv4 & Convolution & $n:384, k:3\times 3, s:1, p:1$ \\\hline
		\rule{0pt}{9pt} conv3 & Convolution & $n:384, k:3\times 3, s:1, p:1$ \\ \hline
		\rule{0pt}{9pt} pool2 & MaxPooling & $k:3\times 3, s:2$ \\ \hline
		\rule{0pt}{9pt} conv2 & Convolution & $n:256, k:5\times 5, s:1, p:2$ \\ \hline
		\rule{0pt}{9pt} pool1 & MaxPooling & $k:3\times 3, s:2$ \\ \hline
		\rule{0pt}{9pt} conv1 & Convolution & $n:96, k:11\times 11, s:4 ,p:0$\\ \hline
		\rule{0pt}{9pt} data & Input & $227\times 227$ binary image\\ \hline
	\end{tabular}
	\label{net_config}
\end{table}

In the CNN training, we use a set of curve-structure image pairs $(I_T, I_i^k)$ with ground-truth labels of true or false matching and train the CNN by minimizing the contrastive loss
\begin{align*} 
l(I_T, I_i^k)&\left\Vert \mathbf{f}(I_T)-\mathbf{f}(I_i^k) \right\Vert^2+ \\ 
&(1-l(I_T, I_i^k)){\rm max}\left(m-\left\Vert \mathbf{f}(I_T)-\mathbf{f}(I_i^k) \right\Vert,0\right)^2, 
\end{align*}
where $l(I_T, I_i^k)=1$ if $(I_T, I_i^k)$ is a true matching and $l(I_T, I_i^k)=0$ otherwise, $m>0$ is a margin for separation, and $\mathbf{f}(\cdot)$ is the learned feature of an input after the ``GAP" layer. 

One issue in training the proposed CNN is the insufficient number of positive training data, i.e., the matched pair of images $(I_T, I_i^k)$. For each sherd, we can only construct one positive training sample by pairing the template with its ground-truth design, cropped at the ground-truth location with the ground-truth orientation. In practice, we only have a small number of sherds with known ground-truth designs and insufficient training data may lead to over-fitting in CNN training. By following previous work on data augmentation, we take the following strategies to increase the number of positive training samples. 1) Given a positive matching sample $(I_T, I_i^k)$, we simultaneously rotate both of them by an angle $\beta$ to construct a new positive matching sample. More specifically, we set the angle $\beta$ to be $0^{\circ}$, $45^{\circ}$, $90^{\circ}$, $135^{\circ}$, $180^{\circ}$, $225^{\circ}$, $270^{\circ}$, and $315^{\circ}$. 2) Given a positive matching sample $(I_T, I_i^k)$, we can simultaneously flip both of them, horizontally or vertically or non-flipping, to construct new positive samples. 3)  We apply FishEye transformation to each positive sample by different exponents (1, 1.25, 1.5 and 1.75) to construct new positive samples. By combing these three strategies, for each sherd with its known ground-truth design, as well as the matched location and orientation, we can construct $8\times 3\times 4 =96$ positive training samples. 

Similarly, in the testing stage, we also apply the same strategies to compute a more reliable CNN-based matching cost. More specifically, given a pair of candidate matching $(I_T, I_i^k)$, we construct 96 candidate matchings using the combined rotation, flipping and FishEye transformation as mentioned above. We then compute the CNN-based matching cost $\psi(\cdot, \cdot)$ for each of these  96 pairs, by feeding the input pair of images to the two sub-networks of the trained CNN, respectively. We find that the final features $\mathbf{f}(\cdot)$ is usually inappropriate for measuring this CNN-based matching cost since the above training is based on binary classification, i.e., the two inputs are either matched or not matched. Instead, in the testing stage, the resulting matching cost is expected to be a real number. Therefore, we define their CNN-based matching cost $\psi(\cdot, \cdot)$ using Euclidean norm. For example, for the original pair $(I_T, I_i^k)$ without any rotation, flipping, and FishEye transformation, this matching cost is computed by
\begin{equation}
	\psi(I_T, I_i^k)= \left\Vert \mathbf{f}(I_T)-\mathbf{f}(I_i^k) \right\Vert^2,
	\label{eq:CNN-matching-cost}
\end{equation}
where $\mathbf{f}(\cdot)$ is the learned feature of an input of the trained CNN as shown in Fig.~\ref{fig:8-network}. For each of the 96 pairs derived from $(I_T, I_i^k)$, we calculate their matching cost $\psi(\cdot, \cdot)$ and denote their average  as $\bar \psi(I_T, I_i^k)$. We perform re-ranking of the candidate matchings according to this averaged matching cost $\bar \psi(\cdot, \cdot)$. The highest ranked candidate matchings, i.e., the ones with the lowest average matching costs, provide the matched designs and the matched location and orientation on these designs. 

The whole algorithm is summarized as shown in Algorithm~\ref{algorithm:flow}.

\begin{algorithm}
\caption{Algorithm for design identification.}
\label{algorithm:flow}
\begin{algorithmic}
\STATE Input: A sherd depth image; designs $I_1,I_2, \cdots, I_N$ 
\STATE Segment input sherd image for curve structure $I_T$ and mask $M_T$ 
\FOR {$i=1$ \textbf{to} $N$}
\FOR {each translation $\bf t$}
\FOR {each rotation $\theta\in[0^{\circ}, 360^{\circ})$ } 
    \STATE  Calculate template matching cost $\phi_i(\mathbf{p},\theta)$ by Eq.~(\ref{eq:template-matching-cost}) 
\ENDFOR	
\ENDFOR
\STATE Select $K$ candidate matchings on design $i$ using ``non-minimum suppression"
\ENDFOR
\FOR {$i=1$ \textbf{to} $N$}
\FOR {$k=1$ \textbf{to} $K$}
\STATE Construct 96 pairs of input from $(I_T, I_i^k)$ by data augmentation
\STATE Compute the average matching cost $\bar \psi(I_T, I_i^k)$
\ENDFOR
\ENDFOR
\STATE Re-rank $NK$ candidate matchings based on $\bar \psi$
\end{algorithmic}
\end{algorithm}

\section{Experiment}
In this section, we conduct experiments to validate the effectiveness of the proposed method. We first quantitatively evaluate the performance of the proposed method in terms of CMC metric and compare it against eight existing methods. Then, we conduct an ablation study to justify the usefulness of both stages in the proposed matching algorithm.

\subsection{Dataset and Settings}

For our study, we collected 600 pottery sherds that were excavated in various archaeological sites located in Southeastern North America. These 600 sherds represent 98 unique paddle designs. Each sherd only displays one design, while the same design may be applied to the surface of multiple sherds. We used a linear array 3D laser scanner, NextEngine, to get the point cloud of a sherd surface with the resolution of 100 points per $mm^2$. Then its depth image is constructed by following the same resolution, i.e., each pixel in the depth image covers an area of $0.01mm^2$. The scanner is placed about 9 inches above the sherd and is perpendicular to the platform where the sherd is seated. The size of the collected depth image ranges from  $130\times100$ pixels to $330\times520$ pixels, while the size of design image ranges from $280\times290$ pixels to $830\times550$ pixels. Archaeologists helped identify the true matching and the true design for each sherd, which we use as ground truth in our experiments. Curve structures are segmented from the depth image of each sherd using our previously developed method as described in Section~\ref{sec:segmentation}.

For computing the matching cost in Eq.~(\ref{eq:CNN-matching-cost}) as shown in Fig.~\ref{fig:8-network}. We set $K=3$, i.e., for each sherd $I_T$, we use template matching to select three candidate matchings on each design. In Stage 2, we randomly select 300 sherds for training the CNN. More specifically, with $K=3$, these 300 training sherds generate $300\times 98\times 3=88,200$ candidate matchings in the form of image pairs in Stage 1. Among them, true matchings are taken as positive training samples while the false matchings are taken as negative training samples. 
We then use the remaining 300 sherds for testing. Similarly with $K=3$, these 300 testing sherds also generate $300\times 98\times 3=88,200$ candidate matchings in the form of image pairs in Stage 1.
The trained CNN is used to re-rank the $98\times 3 = 294$ candidate matchings generated by each testing sherd for identifying the matched design of this testing sherd.

In Stage 2, the proposed re-ranking CNN is initialized by the pre-trained AlexNet model. It is trained by Stochastic Gradient Descent (SGD) with a batch size of 32, momentum of 0.9, and weight decay of 0.0005. The base learning rate is $1\times10^{-4}$, and it decreases slowly in the training process. We set the maximum number of iterations to 50,000 and the margin for separation $m$ to 0.5. 

\subsection{Design Identification Performance}

In this paper, we use the Cumulative Matching Characteristics (CMC) ranking metric to evaluate the design-identification performance. For each sherd $I_T$, we match it against all $N$ designs, generating $NK$ candidate matchings. We then use the average CNN-based matching cost $\bar \psi$ to re-rank these $NK$ candidate matchings. In the re-ranking result, for  the $K$ candidate matchings from the same design,
we only keep the one with the smallest cost $\bar \psi$. This way, we get a re-ranking result of $N$ candidate matchings, each from one different design, with the ascending cost $\bar \psi$. The Rank-$L$ CMC value is the percentage of the $300$ testing sherds with correctly identified designs among the top $L$ candidate matchings in the re-ranking results. By varying $L$ from 1 to $N$, we can draw a CMC curve based on the corresponding Rank-$L$ CMC values. The higher the CMC curve, the better the identification performance.

To evaluate the effectiveness of the proposed method for design identification, we select eight existing matching algorithms for comparison: Template Matching~\cite{brunelli2009template}, Chamfer Matching~\cite{barrow1977parametric}, Shape Context~\cite{belongie2001shape}, Nearest Neighbor~\cite{weber1998quantitative}, pHash~\cite{zauner2010implementation}, Gabor~\cite{manjunath1996texture}, DeepCompare~\cite{zagoruyko2015learning} and MatchNet~\cite{han2015matchnet}. The experiments are conducted on the same testing dataset with 300 sherds. 

Template Matching directly uses the matching cost in Eq.~(\ref{eq:template-matching-cost}) for finding the best matched designs, as well as locations and orientations. In Chamfer Matching, sherd curve structure $I_T$ and each design are first thinned to one-pixel-wide skeleton $U$ and $V$, respectively. Then $U$ is translated and rotated to match $V$ in terms of Chamfer distance
\begin{equation}
\label{eq:chamfer distance}
d_{CM}(U_{\bf T},V) = \frac{1}{|U|}\sum_{{\bf u} \in U_{\bf T}}\min_{{\bf v}\in V}\left \| {\bf u} - {\bf v} \right \|_2, 
\end{equation} 
where $\bf T$ indicates a spatial transform. The Chamfer matching cost is then defined as the minimal $d_{CM}$ over possible $T$'s, including all translations and rotations. Chamfer Matching can be computed efficiently by building a distance map for $V$. 

As in Chamfer Matching, Shape Context also uses the skeletons $U$ and $V$ for matching. Since this is a partial matching and Shape Context is rotation invariant, we slide $U$ over $V$ and calculate the shape-context matching at each location of sliding for best matching locations. We directly use the Shape Context implementation, as well as its matching cost, from the OpenCV package\footnote{\url{https://docs.opencv.org/3.0-beta/modules/shape/doc/shape_distances.html}}
. Nearest Neighbor, pHash, Gabor, DeepCompare and MatchNet use the candidate matchings selected by the Stage 1 of the proposed method and then re-rank the candidate matchings using these matching methods and their respective matching costs. The same CMC ranking metric is then computed for each of them for performance evaluation. Specifically, for Nearest Neighbor, we directly calculate the intensity difference between a pair of inputs as their matching cost. For pHash, we use the radial hash as the hash function, and the parameters follow the setting in~\cite{zauner2010implementation}. pHash was implemented using pHash library\footnote{\url{https://www.phash.org/}}. For Gabor, we construct gabor features using Gabor filter from OpenCV package\footnote{\url{https://docs.opencv.org/3.0-beta/modules/imgproc/doc/filtering.html}}. We set orientation angle  to be $0^{\circ}$, $45^{\circ}$, $90^{\circ}$, $135^{\circ}$, $180^{\circ}, 215^{\circ}$, $270^{\circ}$, and $315^{\circ}$, and wavelength to be $8$, $9$, $10$, and $11$. The matching cost is then defined as Euclidean distance between these Gabor features. For the MatchNet, we employ its original network architecture and training parameters, then fine-tune with the above training dataset on the model trained on ``Yosemite" dataset\footnote{\url{https://github.com/hanxf/matchnet}}. DeepCompare has multiple  networks. We choose the 2-channel deep network introduced in~\cite{zagoruyko2015learning}, and the base model trained on ``Yosemite" dataset\footnote{\url{https://github.com/szagoruyko/cvpr15deepcompare}}. We then fine-tune the model with the canny edge images generated from the ``Yosemite" dataset. 

\begin{figure}[htbp]
	\centering
	\includegraphics[width=1\linewidth]{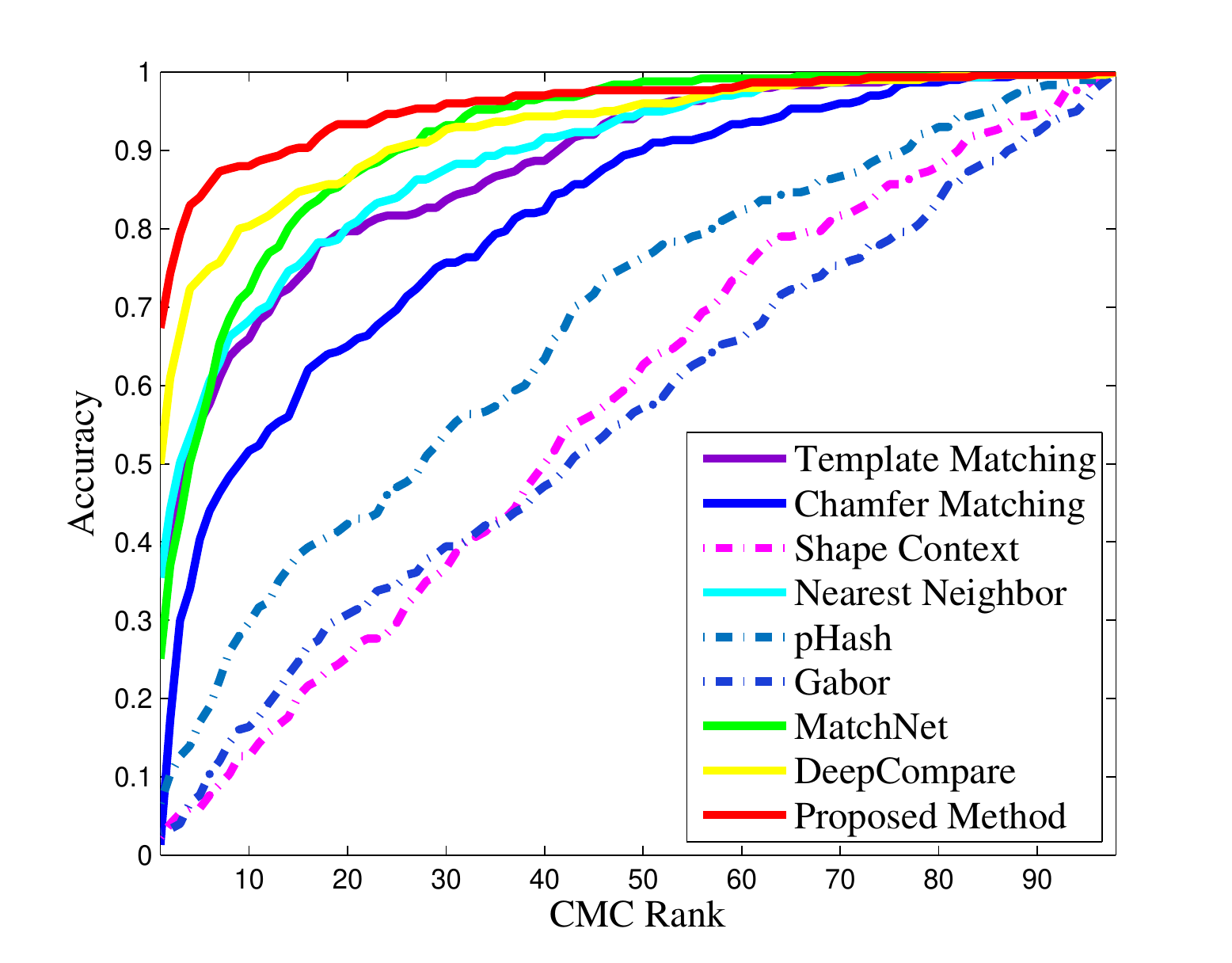}
	\caption{CMC curves of the proposed method and the eight comparison matching methods. }
	\label{fig:9-cmc300}
	\vspace{-0em}
\end{figure}

\begin{figure*}[htbp]
            \begin{center}
            \includegraphics[width=0.8\linewidth]{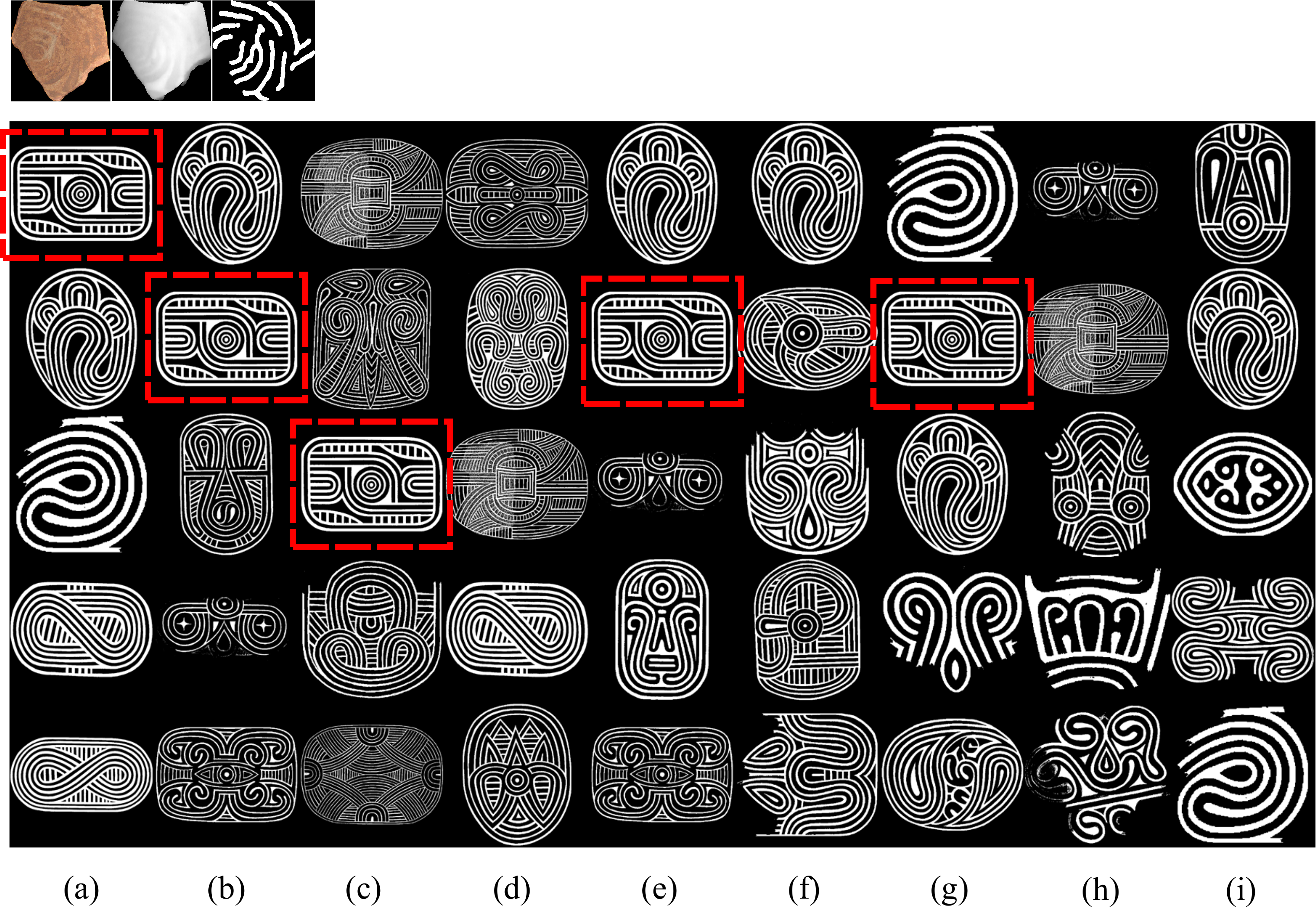}         
	\caption{The top 5 matched designs (from top to the bottom) identified by (a) the proposed method, (b) Template Matching, (c) Nearest Neighbor, (d) MatchNet, (e) DeepCompare, (f) Shape Context, (g) Chamfer matching, (h) Gabor and (i) pHash, respectively. True designs are highlighted in the red box. Original design reproduced with permission, courtesy of
Frankie Snow, South Georgia State College.}
	\label{fig:10-sampleidentificationresult}
	\end{center}
	\vspace{-1.5em}
\end{figure*}

CMC curves of the proposed method and the eight comparison methods are shown in Fig.~\ref{fig:9-cmc300}. We can see  that the proposed method achieves the best CMC performance, and outperforms the second best matching method by 17.3\% on Rank-1 CMC value. Figure~\ref{fig:10-sampleidentificationresult} shows the identification result of the proposed method and the eight comparison methods on a curve structure segmented from a degraded sherd. We can see that, the proposed method matches the true design (in red box) at CMC Rank 1, while the other comparison methods do not. In this figure, we only display the portion masked by the sherd boundary for the top five matched designs. True matchings are highlighted in red boxes.

One interesting finding in Fig.~\ref{fig:9-cmc300} is that Template Matching and Nearest Neighbor methods produce much better results than Chamfer Matching and Shape Context. We believe the major reason is that the thinning of curve structures to skeletons in Chamfer Matching and Shape Context makes them very sensitive to image and structure noise. Both DeepCompare and MatchNet could not well learn the features for the proposed curve-pattern matching with noise and deformations and therefore, using them for re-ranking could not produce satisfactory results. We also found that the handcrafted features, such as Gabor feature cannot well capture spatial information, thus show poor performance in our application.

\subsection{Usefulness of Each Stage}

Intuitively, either of the two stages in the proposed matching algorithm can be replaced by other alternatives or ignored. To justify the usefulness of each stage, we perform several additional experiments, in which we modify or remove one stage of the proposed algorithm, and then check the influence to the identification performance.

\textbf{Removing Stage 1}: Stage 1 of the proposed method uses a highly computationally efficient matching method - template matching - to exhaustively go over the whole search space, consisting of all possible  translations and rotations to align the sherd to a design. To justify its usefulness, we can bypass Stage 1, by directly applying CNN to each solution in the whole search space. Our experiment shows the proposed CNN matching has much higher computational complexity, compared to the template matching in  the proposed Stage 1. For example, just matching one sherd curve structure of size $186 \times 217$ pixels to 3,750 design patches by the proposed CNN take 57.895 seconds. This way, matching such a sherd curve structure to a design of size $326 \times 271 $ pixels over the whole search space takes 7.24 hours on a DELL Precision T7600 workstation with an Intel Xeon E5-2650 CPU, 32GB memory and an nVidia Tesla K20 GPU card. But exhaustive template matching over the same search space takes only 2.178 seconds on the same computer. With the increased number of designs, sherds and the increased image resolutions, it is impossible to run CNN over the whole search space exhaustively.

\begin{figure}[htbp]
	\centering
	\vspace{-1em}
	\includegraphics[width=0.8\linewidth]{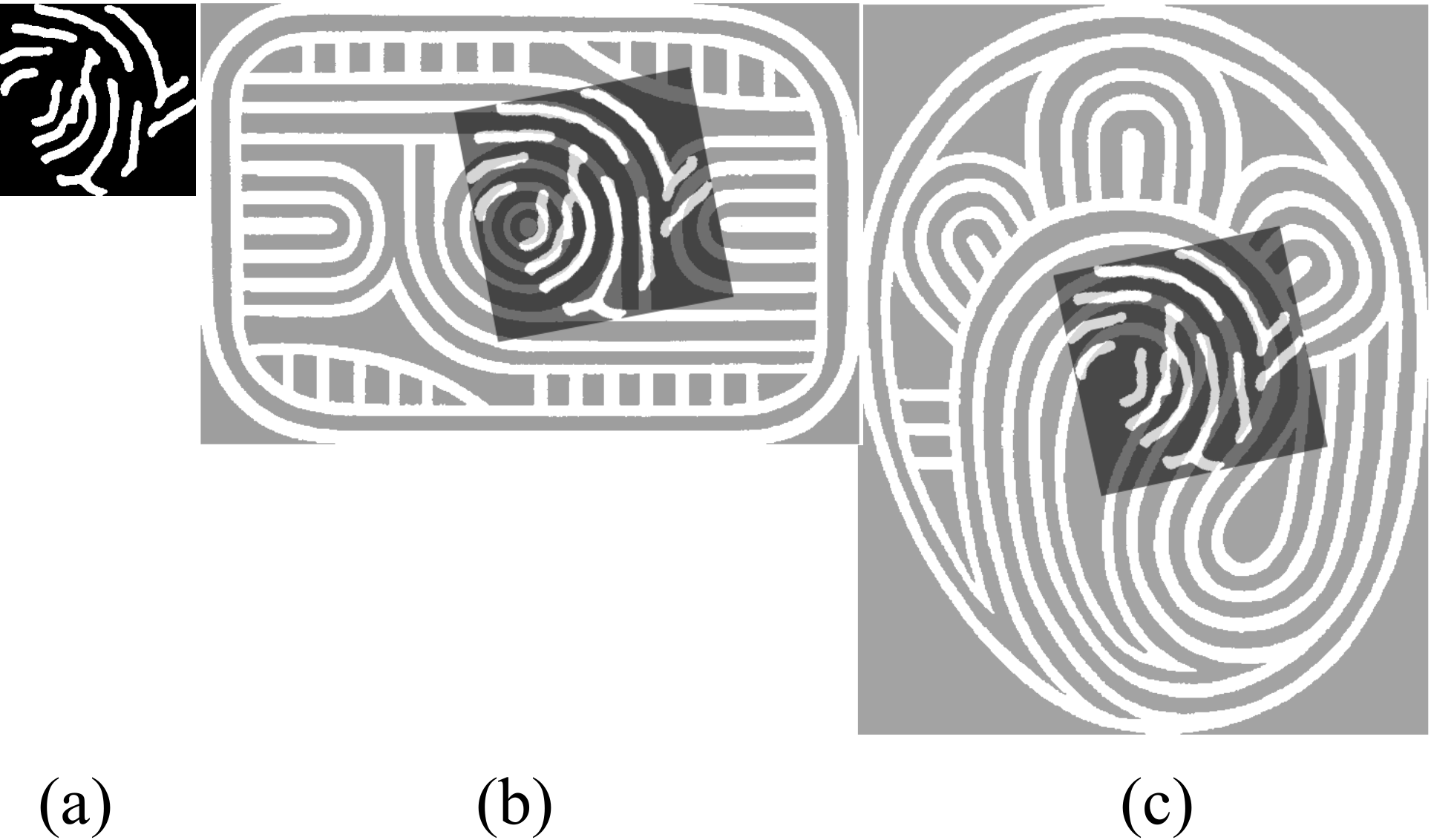}
	\caption{An illustration of the inappropriateness of applying the trained CNN over the whole search space. (a) Curve structure of a sherd, (b) matching to the true location and orientation, with $\bar{\psi}=39.708$, and (c) matching to another location and orientation, with lower $\bar{\psi}=37.512$. Original design reproduced with permission, courtesy of
Frankie Snow, South Georgia State College.}
	\label{fig:11-CNN-matching}
	\vspace{-0.5em}
\end{figure}

Furthermore, we found that the trained CNN may not be able to correctly locate the true locations and orientations on the matched designs when applying it exhaustively over the whole search space, i.e., removing Stage 1. To validate our findings, for each sherd in testing dataset, we randomly select 39 masked design patches located in the range of $(\mathbf{p}\pm9,\theta\pm5^{\circ})$ in each design, while $(\mathbf{p},\theta)$ indicates the location and orientation with the best template matching in each design. We also include masked design patches with the best template matching in each design. We apply the proposed CNN matching on the sherds in training dataset and the above selected design patches, the Rank-1 CMC value drops from 67.3\% to 29.3\%. As shown by an example in Fig.~\ref{fig:11-CNN-matching}, the CNN matching cost $\bar  \psi$ is not the minimum when the sherd is matched to the true location with the true orientation in its true design. We believe this is due to the use of limited training samples for the proposed CNN -- both positive and negative training samples are the candidate matchings selected in Stage 1, and could not represent all possible matchings in the whole search space. To make it work over the whole search space, we have to use a much larger number of training samples. This is difficult since we only have a relatively small number of positive training samples. 

\begin{figure}[htbp]	
	\centering
	\vspace{-1em}
	\includegraphics[width=1\linewidth]{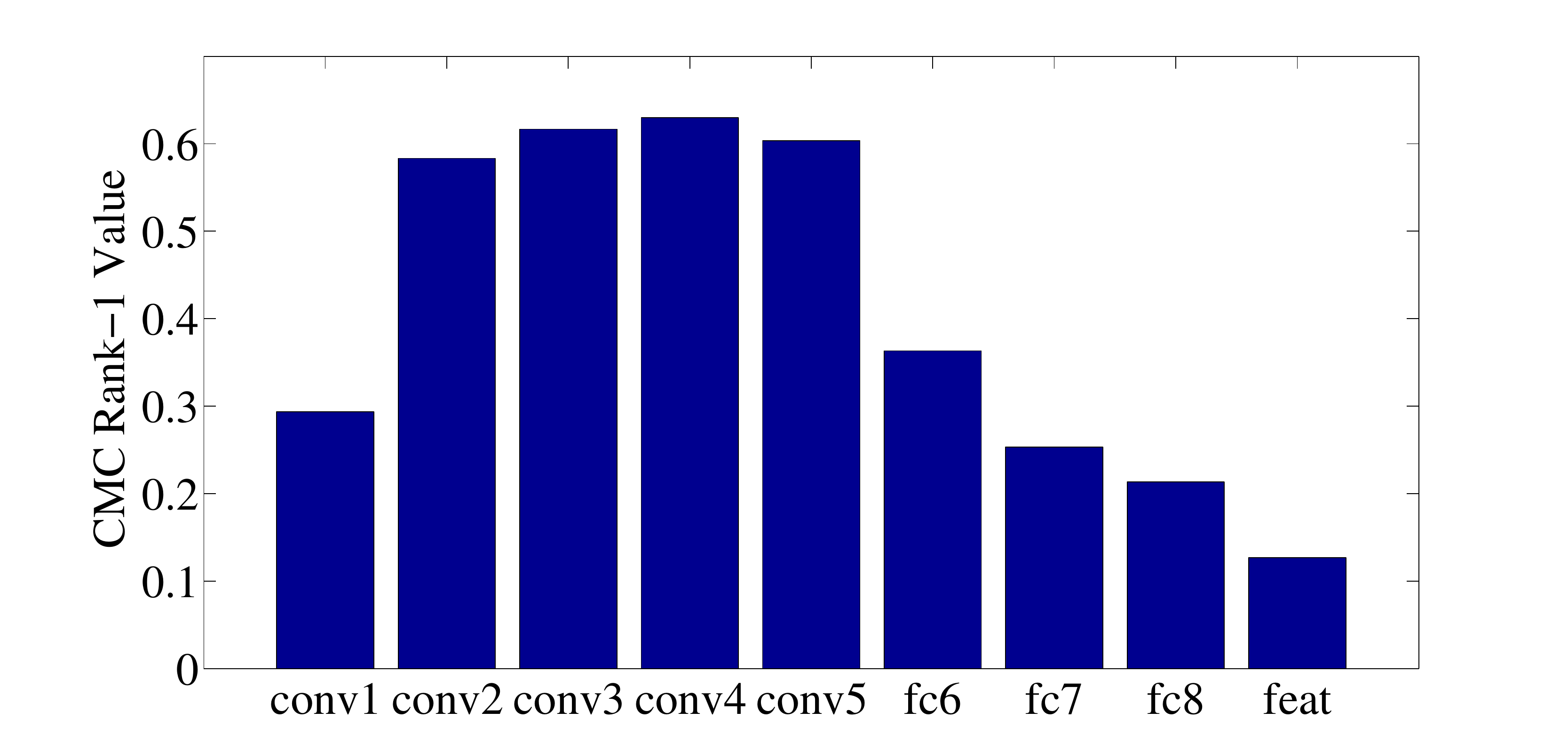}
	\vspace{-1em}
	\caption{Rank-1 CMC values of AlexNet on the 300 testing sherds using learned features from each layer of AlexNet.}
	\label{fig:12-layer-comp}
	\vspace{-0.5em}
\end{figure}

\textbf{Modifying Stage 2}: Stage 2 of the proposed method employs the average CNN-based matching cost $\bar \psi$ to re-rank all the candidate matchings generated in Stage 1. In Stage 2, we truncate AlexNet to ``conv4" layer and add a GAP layer to reduce feature map dimension. Basically, the learned features from each layer of AlexNet, shown in Fig.~\ref{fig:8-network}(b), can be used to calculate the matching cost. However, in practice, we truncate all the layers after ``conv4" because it achieves the optimal identification performance using features from ``conv4" layer. To validate this choice, we conducted comparison experiments by replacing the proposed CNN with AlexNet and utilizing features from every layer of the AlexNet for defining the CNN-based matching cost denoted by Eq.~(\ref{eq:CNN-matching-cost}). The AlexNet was trained using the same training dataset. Note that different from the original AlexNet, we replace the last softmax layer with a ``feat" layer to reduce the feature map dimension, since the original AlexNet was designed for multiple classes, while our application needs a binary classifier. Figure~\ref{fig:12-layer-comp} shows the Rank-1 CMC values when using features from different CNN layers.  

We can see that the highest Rank-1 CMC value is achieved when using the features from ``conv4" layer, not the features from the final ``feat" layer, although the CNN was trained to minimize the loss after the ``feat" layer. The major reason is that the CNN is trained as a binary classifier, while the matching cost is a real value reflecting the similarity between two input images. Different layer can measure the image similarity at different levels of details. From ``conv1" to ``feat", the features become more and more abstracted. However, we desire an image similarity to be measured at a geometric structure level, which might be better reflected at a hidden layer instead of the final layer. Therefore, we believe it is reasonable to construct the proposed CNN by truncating AlexNet to ``conv4" layer. 

In Stage 2, we use data augmentation strategy to improve the testing performance by using average matching cost over 96 pairs of images. We conducted a comparison experiment by removing this data augmentation strategy. Our experiment shows the data augmentation in testing improve Rank-1 CMC value from 60.7\% to 67.3\%. Note that this experiment justifies the use of data augmentation in testing. In the training of CNN, data augmentation is always used and its effectiveness has been verified by many previous works~\cite{krizhevsky2012imagenet,yaeger1997effective,masi2016we}.

\subsection{Running Time}

For identifying the underlying design of a sherd, we need to match the curve structure segmented from the sherd against all 98 designs and Table~\ref{CPU-time} reports the running time for identifying the design of one sherd, averaged over 300 testing sherds. The experiments are conducted on a DELL Precision T7600 workstation with an Intel Xeon E5-2650 CPU, 32GB memory and an nVidia Tesla K20 GPU card.
We can see that the proposed method runs faster than Chamfer Matching and Shape Context because it uses the efficient Template Matching in Stage 1 to reduce the search space. However, it is less efficient than Template Matching due to the additional Stage 2 of re-ranking. Nearest Neighbor, pHash, Gabor, MatchNet and DeepCompare take less running time than the proposed method because they do not consider data augmentation. 

\begin{table}[htbp]
\caption{Average running time of identifying the underlying design of a sherd using the proposed method and eight comparison methods.}
	\centering
	\begin{tabular}{|c |c |} 
		\hline
		\rule{0pt}{10pt} \textbf{Method} & \textbf{Running time (hour)} \\
		\hline\hline
		\rule{0pt}{9pt} Template Matching & 0.969 \\ \hline
		\rule{0pt}{9pt} Chamfer Matching & 29.90 \\ \hline
		\rule{0pt}{9pt} Shape Context & 4.90 \\ \hline
		\rule{0pt}{9pt} Nearest Neighbor & 0.969 \\ \hline
		\rule{0pt}{9pt} pHash &   0.972\\ \hline
		\rule{0pt}{9pt} Gabor &   0.970\\ \hline
		\rule{0pt}{9pt} MatchNet & 0.970  \\\hline
		\rule{0pt}{9pt} DeepCompare & 0.970  \\\hline
		\rule{0pt}{9pt} Proposed & 1.129  \\\hline		
	\end{tabular}
	\label{CPU-time}
	\vspace{-1em}
\end{table}

\section{Conclusion}
In this paper, we explored an important and challenging task in archeology: identifying the underlying curve design on the surface of a highly fragmented and degraded pottery sherd. We developed a new 2-stage template matching algorithm to match the curve structure segmented from a sherd to a set of known designs. In Stage 1, we used a computationally efficient template-matching algorithm to select a small set of candidate matchings on all the designs. In Stage 2, we developed a new CNN-based model to re-rank the candidate matchings for identifying the underlying matched designs. In the experiment, we validated the proposed method by using
600 real sherds together with their corresponding 98 designs from the Woodland Period in Southeastern North America. Comparison to eight existing matching methods verified that the proposed method can achieve a new state-of-the-art performance with reasonable computation time.

\section*{Acknowledgment}

This research was supported by the National Science Foundation Archaeology and Archaeometry Grant Program (1658987), the National Center for Preservation Technology and Training Grants Program (P16AP00373) and University of South Carolina Social Sciences Grant Program. We would like to show our gratitude to Professor Frankie Snow at South Georgia State College for sharing his pearls of wisdom and design images with us during the course of this research. We also thank Dr. Matthew Compton, Curator of the R. M. Bogan Repository at Georgia Southern University for generously sharing his collection, and our colleague Professor Scot Keith for encouraging the pursuit of this research.

\ifCLASSOPTIONcaptionsoff
  \newpage
\fi

\bibliographystyle{IEEEtran}
\bibliography{zhou_bib}

\begin{IEEEbiographynophoto}
{Jun Zhou}
received her BS degree in computer science at Nankai University in China and her MS degree in computer science at Loyola University at Chicago in 1996 and 2000, respectively. She is a PhD candidate in the Department of Computer Science and
Engineering at the University of South Carolina. She works as the director of Research Computing Center at College of Arts and Sciences, University of South Carolina. Her interests include document image processing and cultural heritage object image processing.
\end{IEEEbiographynophoto}
\begin{IEEEbiographynophoto}
{Yuhang Lu}
received his B.E. degree from Chengdu University of Technology in 2013 and his M.E. degree from Wuhan University in 2015. He is currently a Ph.D. student in the Department of Computer Science and Engineering at the University of South Carolina. His research interests include computer vision, machine learning and cultural heritage object image processing.
\end{IEEEbiographynophoto}
\begin{IEEEbiographynophoto}
{Kang Zheng}
received his B.E. degree in Electrical Engineering from Harbin Institute of Technology in 2012. He is currently a Ph.D. candidate with Department of Computer Science and Engineering, University of South Carolina. His research interests include computer vision, image processing and deep learning.
\end{IEEEbiographynophoto}
\begin{IEEEbiographynophoto}
{Karen Smith}
obtained her PhD from the University of Missouri in
2009. She is a southeastern archaeologist with a background in Woodland period and plantation-era research and archaeological curation. She works as the director of Applied Research Division, South Carolina Institute of Archaeology and Anthropology at the University of South Carolina. Her interests include Woodland period and plantation research and archaeological data analysis.
\end{IEEEbiographynophoto}
\begin{IEEEbiographynophoto}
{Colin Wilder}
obtained his BA degree in philosophy from Yale University
and his PhD in German history from the University of Chicago in
2010. He is the associate director in the Center for Digital Humanities
and the director of the Republic of Literature and of the Dirty History
Metacrawler at the University of South Carolina. His research focuses
on the development of the ideas of liberty and equality in German and
broader European history in the Early Modern period.
\end{IEEEbiographynophoto}
\begin{IEEEbiographynophoto}
{Song Wang}
received his PhD degree in electrical and computer engineering
from the University of Illinois at Urbana-Champaign in 2002.
He is currently a professor in the Department of Computer Science
and Engineering, University of South Carolina. His current research
interests include computer vision, image processing, and machine
learning. He serves as an associate editor of Pattern Recognition
Letters. He is a senior member of IEEE and a member of the IEEE
Computer Society.
\end{IEEEbiographynophoto}

\end{document}